\documentclass[11pt,a4paper]{article}
\usepackage[hyperref]{naaclhlt2019}
\usepackage{times}
\usepackage{latexsym}

\usepackage{url}

\usepackage{graphicx}
\usepackage{amsmath}
\usepackage{color}
\usepackage{algorithm}
\usepackage[noend]{algpseudocode}

\usepackage{capt-of}
\usepackage{varwidth}
\newsavebox\tmpbox

\aclfinalcopy 
\def\aclpaperid{724} 

\title{Playing Text-Adventure Games with\\Graph-Based Deep Reinforcement Learning}

\author{Prithviraj Ammanabrolu \\
  School of Interactive Computing \\
  Georgia Institute of Technology \\
  Atlanta, GA \\
  {\tt raj.ammanabrolu@gatech.edu} \\\And
  Mark O. Riedl \\
  School of Interactive Computing \\
  Georgia Institute of Technology \\
  Atlanta, GA \\
  {\tt riedl@cc.gatech.edu} \\}

\date{}

\begin{document}
\maketitle
\begin{abstract}

Text-based adventure games provide a platform on which to explore reinforcement learning in the context of a combinatorial action space, such as natural language.
We present a deep reinforcement learning architecture that represents the game state as a knowledge graph which is learned during exploration.
This graph is used to prune the action space, enabling more efficient exploration.
The question of which action to take can be reduced to a question-answering task, a form of transfer learning that pre-trains certain parts of our architecture.
In experiments using the TextWorld framework, we show that our proposed technique can learn a control policy faster than baseline alternatives.
We have also open-sourced our code at https://github.com/rajammanabrolu/KG-DQN.

\end{abstract}

\section{Introduction}
Natural language communication can be used to affect change in the real world.
Text adventure games, in which players must make sense of the world through text descriptions and declare actions through natural language, can provide a stepping stone toward more real-world environments where agents must communicate to understand the state of the world and indirectly affect change in the world.
Text adventure games are also useful for developing and testing reinforcement learning algorithms that must deal with the partial observability of the world \citep{Narasimhan2015,He2015}.

In text adventure games, the agent receives an incomplete textual description of the current state of the world.
From this information, and previous interactions with the world, a player must determine the next best action to take to achieve some quest or goal.
The player must then compose a textual description of the action they intend to make and receive textual feedback of the effects of the action.
Formally, a text-based game is a partially observable Markov decision process (POMDP), represented as a 7-tuple of $\langle S,T,A,\Omega , O,R, \gamma\rangle$ representing the set of environment states, conditional transition probabilities between states, words used to compose text commands, observations, observation conditional probabilities, reward function, and the discount factor respectively~\citep{Cote2018}.

In text-based games, 
the agent never has access to the true underlying world state and has to reason about how to act in the world based only on the textual observations. 
Additionally, the agent's actions must be expressed through natural language commands, ensuring that the action space is combinatorially large.
Thus, text-based games pose a different set of challenges than traditional video games.
Text-based games require a greater understanding of previous context to be able to explore the state-action space more effectively.
Such games have historically proven to be difficult to play for AI agents, and the more complex variants such as \textit{Zork} still remain firmly out of the reach of existing approaches. 

We introduce three contributions to text-based game playing to deal with the combinatorially large state and action spaces.
%
First, we show that a state representation in the form of a {\em knowledge graph} gives us the ability to effectively prune an action space. 
A knowledge graph captures the relationships between entities as a directed graph.
The knowledge graph provides a persistent memory of the world over time and enables the agent to have a prior notion of what actions it should not take at a particular stage of the game.

Our second contribution is a deep reinforcement learning architecture, {\em Knowledge Graph DQN} (KG-DQN), that effectively uses this state representation to estimate the $Q$-value for a state-action pair.
This architecture leverages recent advances in graph embedding and attention techniques \citep{Guan2018,velickovic2018graph} to learn which portions of the graph to pay attention to given an input state description in addition to having a mechanism that allows for natural language action inputs.
Finally, we take initial steps toward framing the POMDP as a question-answering (QA) problem wherein a knowledge-graph can be used to not only prune actions but to answer the question of what action is most appropriate. 
Previous work has shown that many NLP tasks can be framed as instances of question-answering and that we can transfer knowledge between these tasks \citep{Mccann2017}. 
We show how pre-training certain parts of our KG-DQN network using existing QA methods improves performance and allows knowledge to be transferred from different games.

We provide results on ablative experiments comparing our knowledge-graph based approach approaches to strong baselines.
Results show that incorporating a knowledge-graph into a reinforcement learning agent results in converges to the highest reward more than $40\%$ faster than the best baseline.
With pre-training using a question-answering paradigm, we achieve this fast convergence rate while also achieving high quality quest solutions as measured by the number of steps required to complete the quests.

\section{Related Work}

A growing body of research has explored the challenges associated with text-based games \citep{weston10,Narasimhan2015,He2015,fulda-ijcai2017,Haroush2018,Cote2018,tao18}. 
\citet{Narasimhan2015} attempts to solve parser-based text games by encoding the observations using an LSTM. 
This encoding vector is then used by an action scoring network that determines the scores for the action verb and each of the corresponding argument objects. 
The two scores are then averaged to determine $Q$-value for the state-action pair. 
\citet{He2015} present the Deep Reinforcement Relevance Network (DRRN) which uses two separate deep neural networks to encode the state and actions. 
The $Q$-value for a state-action pair is then computed by a pairwise interaction function between the two encoded representations. %
Both of these methods are not conditioned on previous observations and so are at a disadvantage when dealing with complex partially observable games. 
Additionally, neither of these approaches prune the action space and so end up wasting trials exploring state-action pairs that are likely to have low $Q$-values, likely leading to slower convergence times for combinatorially large action spaces.

\citet{Haroush2018} introduce the Action Eliminating Network (AEN) that attempts to restrict the actions in each state to the top-$k$ most likely ones, using the emulator's feedback. 
The network learns which actions should not be taken given a particular state. 
Their work shows that reducing the size of the action space allows for more effective exploration, leading to better performance.
Their network is also not conditioned on previous observations.

Knowledge graphs have been demonstrated to improve natural language understanding in other domains outside of text adventure games. 
For example, \citet{Guan2018} use commonsense knowledge graphs such as {\em ConceptNet} \citep{Speer2012} to significantly improve the ability of neural networks to predict the end of a story. 
They represent the graph in terms of a knowledge context vector using features from {\em ConceptNet} and graph attention \citep{velickovic2018graph}. 
The state representation that we have chosen as well as our method of action pruning builds on the strengths of existing approaches while simultaneously avoiding the shortcomings of ineffective exploration and lack of long-term context.

\section{Knowledge Graph DQN}

In this section we introduce our knowledge graph representation, action pruning and deep $Q$-network architecture.  

\subsection{Knowledge Graph Representation}
\label{sec:graphs}

\begin{figure*}
  \centering
  \includegraphics[width=\linewidth]{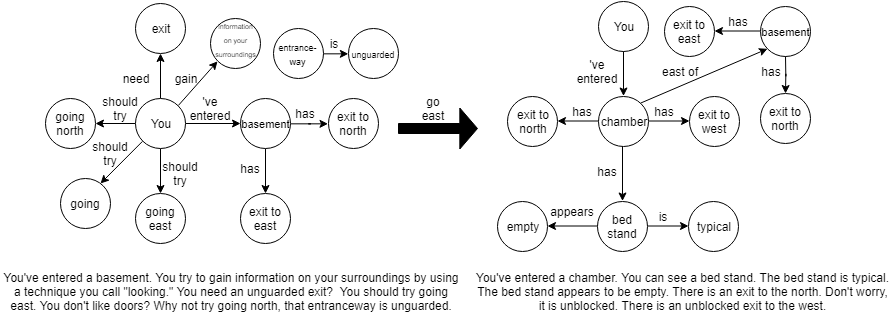}
  \caption{Graph state update example given two observations}
  \label{fig:graphs}
\end{figure*}

In our approach, our agent learns a knowledge graph, stored as a set of RDF triples, i.e. 3-tuples of $\langle subject, relation, object \rangle$. 
These triples are extracted from the observations using Stanford's Open Information Extraction (OpenIE) \citep{Angeli2015}.
OpenIE is not optimized to the regularities of text adventure games and there are a lot of relations that can be inferred from the typical structure of descriptive texts.
For example, from a phrase such as ``There is an exit to the north'' one can infer a {\em has} relation between the current location and the direction of the exit.
These additional rules fill in the information not provided by OpenIE.
The resultant knowledge graph gives the agent what essentially amounts to a mental map of the game world. 

The knowledge graph is updated after every agent action (see Figure~\ref{fig:graphs}).
The update rules are defined such that there are portions of the graph offering short and long-term context. 
A special node---designated ``you''---represents the agent and relations out of this node are updated after every action with the exception of relations denoting the agent's inventory.
Other relations persist after each action.
%
We intend for the update rules to be applied to text-based games in different domains and so only hand-craft a minimal set of rules that we believe apply generally. They are:
\begin{itemize}
    \item Linking the current room type (e.g. ``basement'', ``chamber') to the items found in the room with the relation ``has'',
        e.g. $\langle chamber, has, bed$ $stand\rangle$
    \item Extracting information regarding entrances and exits and linking them to the current room,
        e.g. $\langle basement, has, exit$ $to$ $north \rangle$
    \item Removing all relations relating to the ``you'' node with the exception of inventory every action,
        e.g. $\langle you, have, cubical$ $key\rangle$
    \item Linking rooms with directions based on the action taken to move between the rooms,
        e.g. $\langle chamber, east$ $of, basement\rangle$ after the action ``go east'' is taken to go from the basement to the chamber
\end{itemize}
All other RDF triples generated are taken from OpenIE.

\subsection{Action Pruning}
\label{sec:pruning}

The number of actions available to an agent in a text adventure game can be quite large: $A=\mathcal{O}(|V| \times |O|^2)$ where $V$ is the number of action verbs, and $O$ is the number of distinct objects in the world that the agent can interact with, assuming that verbs can take two arguments.
Some actions, such as movement, inspecting inventory, or observing the room, do not have arguments.

The knowledge graph is used to prune the combinatorially large space of possible actions available to the agent as follows.
Given the current state graph representation $G_t$, the action space is pruned by ranking the full set of actions and selecting the top-$k$.
Our action scoring function is: 
\begin{itemize}
\item +1 for each object in the action that is present in the graph; and
\item +1 if there exists a valid directed path between the two objects in the graph.
\end{itemize}
We assume that each action has at most two objects (for example inserting a key in a lock).

\subsection{Model Architecture and Training}
\label{sec:architecutre}

Following \citet{Narasimhan2015}, all actions $A$ that will be accepted by the game's parser are available to the agent at all times. 
When playing the game, the agent chooses an action and receives an observation $o_t$ from the simulator, which is a textual description of current game state.
The state graph $G_t$ is updated according to the given observation, as described in Section~\ref{sec:graphs}.

We use the $Q$-Learning technique \citep{Watkins1992} to learn a control policy $\pi (a_t|s_t)$, $a_t \in A$, 
which gives us the probability of taking action $a_t$ given the current state $s_t$. The policy is determined by the $Q$-value of a particular state-action pair, which is updated using the Bellman equation \citep{Sutton2018}:
\begin{equation}
\begin{aligned}
    Q_{t+1}(s_{t+1}, &a_{t+1}) =\\
   & E[r_{t+1} + \gamma \max_{a \in A_{t}} Q_t(s, a)|s_t,a_t]
\end{aligned}
\end{equation}
where $\gamma$ refers to the discount factor and $r_{t+1}$ is the observed reward. 
The policy is thus to take the action that maximizes the $Q$-value in a particular state, which will correspond to the action that maximizes the reward expectation given that the agent has taken action $a_t$ at the current state $s_t$ and followed the policy $\pi (a|s)$ after. 

\begin{figure*}
  \centering
  \includegraphics[width=\linewidth]{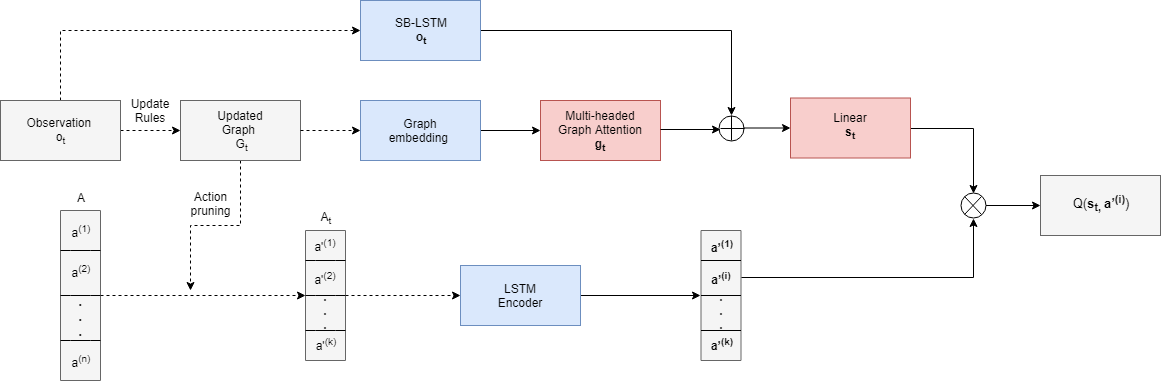}
  \caption{KG-DQN architecture, blue shading indicates components that can be pre-trained and red indicates no pre-training. The solid lines indicate gradient flow for learnable components.}
  \label{fig:architecture}
\end{figure*}

The architecture in Figure~\ref{fig:architecture} is responsible for computing the representations for both the state $s_t$ and the actions $a^{(i)} \in A$ and coming to an estimation of the $Q$-value for a particular state and action. 
%
During the forward activation, the agent uses the observation to update the graph $G_t$ using the rules outlined in Section~\ref{sec:pruning}. 

The graph is then embedded into a single vector $\mathbf{g_t}$. 
We use Graph Attention \citep{velickovic2018graph} with an attention mechanism similar to that described in \citet{Bahdanau2014NeuralMT}.
Formally, the Multi-headed Graph Attention component receives a set of node features $H=\{\mathbf{h_1}, \mathbf{h_2}, \dots, \mathbf{h_N}\}$, $\mathbf{h_i} \in \rm I\!R^F$, where $N$ is the number of nodes and $F$ the number of features in each node, and the adjacency matrix of $G_t$.
Each of the node features consist of the averaged word embeddings for the tokens in that node, as determined by the preceding graph embedding layer.
The attention mechanism is set up using self-attention on the nodes after a learnable linear transformation $W \in \rm I\!R^{2F \times F}$ applied to all the node features:
\begin{equation}
    e_{ij} = LeakyReLU(\mathbf{p} \cdot W(\mathbf{h_i} \oplus \mathbf{h_j}))
\end{equation}
where $\mathbf{p} \in \rm I\!R^{2F}$ is a learnable parameter. The attention coefficients $\alpha _{ij}$ are then computed by normalizing over the choices of $k \in \mathcal{N}$ using the softmax function. Here $\mathcal{N}$ refers to the neighborhood in which we compute the attention coefficients.
This is determined by the adjacency matrix for $G_t$ and consists of all third-order neighbors of a particular node.
\begin{equation}
    \alpha _{ij} = \frac{exp(e_{ij})}{\sum_{k \in \mathcal{N}} exp(e_{ik})}
\end{equation}
Multi-head attention is then used, calculating multiple independent attention coefficients. The resulting features are then concatenated and passed into a linear layer to determine $\mathbf{g_t}$:
\begin{equation}
    \mathbf{g_t} = f(W_g(\Vert_{k=1}^K \sigma (\sum_{j \in \mathcal{N}} \alpha_{ij}^{(k)}\mathbf{W}^{(k)}\mathbf{h}_j)) + b_g)
\end{equation}
where $k$ refers to the parameters of the $k^{th}$ independent attention mechanism, $W_g$ and $b_g$ the weights and biases of this component's output linear layer, and $\Vert$ represents concatenation.

Simultaneously, an encoded representation of the observation $\mathbf{o_t}$ is computed using a Sliding Bidirectional LSTM (SB-LSTM).
The final state representation $\mathbf{s_t}$ is computed as:
\begin{equation}
    \mathbf{s_t} = f(W_l(\mathbf{g_t \oplus o_t}) + b_l)
\end{equation}
where $W_l,b_l$ represent the final linear layer's weights and biases and $\mathbf{o_t}$ is the result of encoding the observation with the SB-LSTM. 

The entire set of possible actions $A$ is pruned by scoring each $a \in A$ according to the mechanism previously described using the newly updated $G_{t+1}$. 
We then embed and encode all of these action strings using an LSTM encoder \citep{sutskever2014}.
The dashed lines in Figure~\ref{fig:architecture} denotes non-differentiable processes.

The final $Q$-value for a state-action pair is:
\begin{equation}
\label{eq:qsa}
    Q(\mathbf{s_t}, \mathbf{a_t}) = \mathbf{s_t}\cdot\mathbf{a_t}
\end{equation}
This method of separately computing the representations for the state and action is similar to the approach taken in the DRRN~\citep{He2015}.

We train the network using experience replay \citep{Lin1993} with prioritized sampling (cf., \cite{Moore1993}) and a modified version of the $\epsilon$-greedy algorithm \citep{Sutton2018} that we call the $\epsilon _1,\epsilon _2$-greedy learning algorithm.
The experience replay strategy finds paths in the game, 
which are then stored as transition tuples in a experience replay buffer $D$.
The $\epsilon _1,\epsilon _2$-greedy algorithm explores by choosing actions randomly from $A$ with probability $\epsilon_1$ and from $A_t$ with a probability $\epsilon_2$.
The second threshold is needed to account for situations where an action must be chosen to advance the quest for which the agent has no prior in $G_t$.
That is, action pruning may remove actions essential to quest completion because those actions involve combinations of entities that have not been encountered before.

We then sample a mini-batch of transition tuples consisting of $\langle\mathbf{s_k}, \mathbf{a_k}, r_{k+1}, \mathbf{s_{k+1}}, \mathbf{A_{k+1}}, p_k\rangle$ from $D$ and compute the temporal difference loss as:
\begin{equation}
\begin{aligned}
    L(\theta) =& r_{k+1} +\\ 
    &\gamma \max_{\mathbf{a} \in \mathbf{A_{k+1}}}Q(\mathbf{s_t}, \mathbf{a}; \theta) - Q(\mathbf{s_t}, \mathbf{a_t}; \theta)
\end{aligned}
\end{equation}
Replay sampling from $D$ is done by sampling a fraction $\rho$ from transition tuples with a positive reward and $1-\rho$ from the rest.
As shown in \cite{Narasimhan2015}, prioritized sampling from experiences with a positive reward helps the deep $Q$-network more easily find the sparse set of transitions that advance the game. 
The exact training mechanism is described in Algorithm 1.

\section{Game Play as Question Answering}

Previous work has shown that many NLP tasks can be framed as instances of question-answering and that in doing so, one can transfer knowledge between these tasks \citep{Mccann2017}. 
In the abstract, an agent playing a text adventure game can be thought of as continuously asking the question ``What is the right action to perform in this situation?'' 
When appropriately trained, the agent may be able to answer the question for itself and select a good next move to execute.
Treating the problem as question-answering will not replace the need for exploration in text-adventure games.
However, we hypothesize that it will cut down on the amount of exploration needed during testing time, theoretically allowing it to complete quests faster;
one of the challenges of text adventure games is that the quests are puzzles and even after training, execution of the policy requires a significant amount of exploration.

To teach the agent to answer the question of what action is best to take given an observation, we use an offline, pre-training approach.
%
The data for the pre-training approach is generated using an oracle, an agent capable of finishing a game perfectly in the least number of steps possible.
Specifically, the agent knows exactly what action to take given the state observation in order to advance the game in the most optimal manner possible. 
Through this process, we generate a set of traces consisting of state observations and actions such that the state observation provides the context for the implicit question of "What action should be taken?" and the oracle's correct action is the answer.
We then use the DrQA~\citep{chen2017reading} question-answering technique to train a paired question encoder and an answer encoder that together predict the answer (action) from the question (text observation).
The weights from the SB-LSTM in the document encoder in the DrQA system are then used to initialize the weights of the SB-LSTM.
Similarly, embedding layers of both the graph and the LSTM action encoder are initialized with the weights from the embedding layer of same document encoder. 
Since the DrQA embedding layers are initialized with GloVe, we are transferring word embeddings that are tuned during the training of the QA architecture.

The game traces used to train the question-answering come from a set of games of the same domain but have different specific configurations of the environment and different quests.
We use the TextWorld framework~\cite{Cote2018}, which uses a grammar to generate random worlds and quests.
The types of rooms are the same, but their relative spatial configuration, the types of objects, and the specific sequence of actions needed to complete the quest are different each time.
This means that the agent cannot simply memorize quests.
For pre-training to work, the agent must develop a general question-answering competence that can transfer to new quests.
Our approach to question-answering in the context of text adventure game playing thus represents a form of transfer learning.




\begin{algorithm*}[tb]
\caption{$\epsilon _1,\epsilon _2$-greedy learning algorithm for KG-DQN}
\scriptsize
\begin{algorithmic}[1]
\For{episode=1 to $M$}
    \State Initialize action dictionary $A$ and graph $G_0$
    \State Reset the game simulator
    \State Read initial observation $o_1$
    \State $G_1 \gets updateGraph(G_0, o_1)$; $A_1 \gets pruneActions(A, G_0)$ \Comment{Section \ref{sec:pruning}}
    \For{step $t$=1 to $T$}
        \If {$random()<\epsilon_1$}
            \If{$random()<\epsilon_2$}
                \State Select random action $a_t \in A$
            \Else
                \State Select random action $a_t \in A_t$
            \EndIf
        \Else
            \State Compute $Q(\mathbf{s_t}, \mathbf{a^{(i)}};\theta)$ for $a^{(i)} \in A$ for network parameters $\theta$ \Comment{Section \ref{sec:architecutre}, Eq. 
            \ref{eq:qsa}}
            \State Select $a_t$ based on $\pi (a|s_t)$
        \EndIf
        \State Execute action $a_t$ in the simulator and observe reward $r_t$
        \State Receive next observation $o{t+1}$
        \State $G_{t+1} \gets updateGraph(G_{t}, o_{t+1})$; $A_{t+1} \gets pruneActions(A, G_{t+1})$ \Comment{Section \ref{sec:graphs}}
        \State Compute $\mathbf{s_{t+1}}$ and $\mathbf{A_{t+1}}=\{\mathbf{a'^{(i)}}$ for all $a'^{(i)}\in A\}$ \Comment{Section \ref{sec:architecutre}}
        \State Set priority $p_t=1$ if $r_t>0$, else $p_t=0$
        \State Store transition ($\mathbf{s_t}, \mathbf{a_t}, r_t, \mathbf{s_{t+1}}, \mathbf{A_{t+1}}, p_t$) in replay buffer $D$
        \State Sample mini-batch of transitions ($\mathbf{s_k}, \mathbf{a_k}, r_k, \mathbf{s_{k+1}}, \mathbf{A_{k+1}}, p_k$) from $D$, with fraction $\rho$ having $p_k=1$
        \State Set $y_k=r_k + \gamma \max_{\mathbf{a} \in \mathbf{A_{k+1}}}Q(\mathbf{s_t}, \mathbf{a}; \theta)$, or $y_k=r_k$ if $s_{k+1}$ is terminal
        \State Perform gradient descent step on loss function $L(\theta)=(y_k - Q(\mathbf{s_t}, \mathbf{a_t}; \theta))^2$
    \EndFor
\EndFor
\end{algorithmic}
\end{algorithm*}

\section{Experiments}

\begin{table}[tb]
\caption{Generated game details.}
\footnotesize
\begin{tabular}{lll}
\hline
                                  & {\bf Small} & {\bf Large} \\ \hline
Rooms                               &   10    & 20      \\ 
Total objects                       &   20    &    40   \\ 
Quest length                        &    5   &  10     \\ 
\hline
Branching factor                    & 143 & 562 \\
Vocab size                          & 746 & 819 \\
Average words per obs.       &  67.5 &    94.0   \\
Average new RDF triples per obs. &   7.2    & 10.5  \\ \hline
\end{tabular}
\label{table:games}
\end{table}

We conducted experiments in the TextWorld framework \citep{Cote2018} using their ``home'' theme. 
TextWorld uses a grammar to randomly generate game worlds and quests with given parameters.
Games generated with TextWorld start with a zero-th observation that gives instructions for the quest; we do not allow our agent to access this information.
The TextWorld API also provides a list of {\em admissible} actions at each state---the actions that can be performed based on the objects that are present.
We do not allow our agent to access the admissible actions.

We generated two sets of games with different random seeds, representing different game difficulties, which we denote as {\em small} and {\em large}. 
Small games have ten rooms and quests of length five and
large games have twenty rooms and quests of length ten.
Statistics on the games are given in Table~\ref{table:games}. 
Quest length refers to the number of actions that the agent is required to perform in order to finish the quest; more actions are typically necessary to move around the environment and find the objects that need to be interacted with. 
The branching factor is the size of the action set $A$ for that particular game.

The reward function provided by TextWorld is as follows: +1 for each action taken that moves the agent closer to finishing the quest; -1 for each action taken that extends the minimum number of steps needed to finish the quest from the current stage; 0 for all other situations. 
The maximum achievable reward for the small and large sets of games are 5 and 10 respectively.
This allows for a large amount of variance in quest quality---as measured by steps to complete the quest---that receives maximum reward.

The following procedure for pre-training was done separately for each set of games.
Pre-training of the SB-LSTM within the question-answering architecture is conducted by generating 200 games from the same TextWorld theme.
The QA system was then trained on data from walkthroughs of a randomly-chosen subset of 160 of these generated games, tuned on a dev set of 20 games, and evaluated on the held-out set of 20 games.
%
Table~\ref{table:pretraining} provides details on the Exact Match (EM), precision, recall, and F1 scores of the QA system after training for the small and large sets of games.
Precision, recall, and F1 scores are calculated by counting the number of tokens between the predicted answer and ground truth.
An Exact Match is when the entire predicted answer matches with the ground truth.
This score is used to tune the model based on the dev set of games.
\begin{table}[tb]
    \caption{Pre-training accuracy.}
    \footnotesize
    \centering
    \begin{tabular}{lllll}
    \hline
      & \textbf{EM}    & \textbf{Precision} & \textbf{Recall} & \textbf{F1}    \\ \hline
    Small & 46.20  & 56.57 & 63.38 & 57.94 \\
    Large & 34.13 & 52.53 & 64.72 & 55.06 \\\hline
\end{tabular}
\label{table:pretraining}
\end{table}

A random game was chosen from the test-set of games and used as the environment for the agent to train its deep $Q$-network on.
Thus, at no time did the QA system see the final testing game prior to the training of the KG-DQN network.

We compare our technique to three baselines: 

\begin{itemize}
\item
Random command, which samples from the list of admissible actions returned by the TextWorld simulator at each step.
\item
LSTM-DQN, developed by Narasimhan et al.~(\citeyear{Narasimhan2015}).
\item
Bag-of-Words DQN, which uses a bag-of-words encoding with a multi-layer feed forward network instead of an LSTM.
\end{itemize}
To achieve the most competitive baselines, we used a randomized grid search to choose the best hyperparameters (e.g., hidden state size, $\gamma$, $\rho$, final $\epsilon$, update frequency, learning rate, replay buffer size) for the BOW-DQN and LSTM-DQN baselines.
%

We tested three versions of our KG-DQN: 
\begin{enumerate}
\item 
Un-pruned actions with pre-training
\item 
Pruned actions without pre-training 
\item 
Pruned actions with pre-training (full)
\end{enumerate}
Our models use 50-dimensional word embeddings, 2 heads on the graph attention layers, mini-batch size of 16, and perform a gradient descent update every 5 steps taken by the agent. 

All models are evaluated by observing the 
(a)~time to reward convergence, 
and 
(b)~the average number of steps required for the agent to finish the game with $\epsilon = 0.1$ over 5 episodes after training has completed.
Following \citet{Narasimhan2015} we set $\epsilon$ to a non-zero value because text adventure games, by nature, require exploration to complete the quests.
All results are reported based on multiple independent trials.
For the large set of games, we only perform experiments on the best performing models found in the small set of games.
Also note that for experiments on large games, we do not display the entire learning curve for the LSTM-DQN baseline, as it converges significantly more slowly than KG-DQN.
We run each experiment 5 times and average the results.

Additionally, human performance on the both the games was measured by counting the number of steps taken to finish the game, with and without instructions on the exact quest.
We modified Textworld to give the human players reward feedback in the form of a score, the reward function itself is identical to that received by the deep reinforcement learning agents.
In one variation of this experiment, the human was given instructions on the potential sequence of steps that are required to finish the game in addition to the reward in the form of a score and in the other variation, the human received no instructions.

\section{Results and Discussion}

Recall that the number of steps required to finish the game for the oracle agent is 5 and 10 for the small and large maps respectively.
It is impossible to achieve this ideal performance due to the structure of the quest.
The player needs to interact with objects and explore the environment in order to figure out the exact sequence of actions required to finish the quest.
To help benchmark our agent's performance, we observed people unaffiliated with the research playing through the same TextWorld ``home'' quests as the other models.
Those who did not receive instructions on how to finish the quest never finished a single quest and gave up after an average of 184 steps on the small map and an average of 190 steps on the large map.
When given instructions, human players completed the quest on the large map in an average of 23 steps, finishing the game with the maximum reward possible.
Also note that none of the deep reinforcement learning agents received instructions.

\begin{figure}[tb]
    \centering
    \includegraphics[width=1\linewidth]{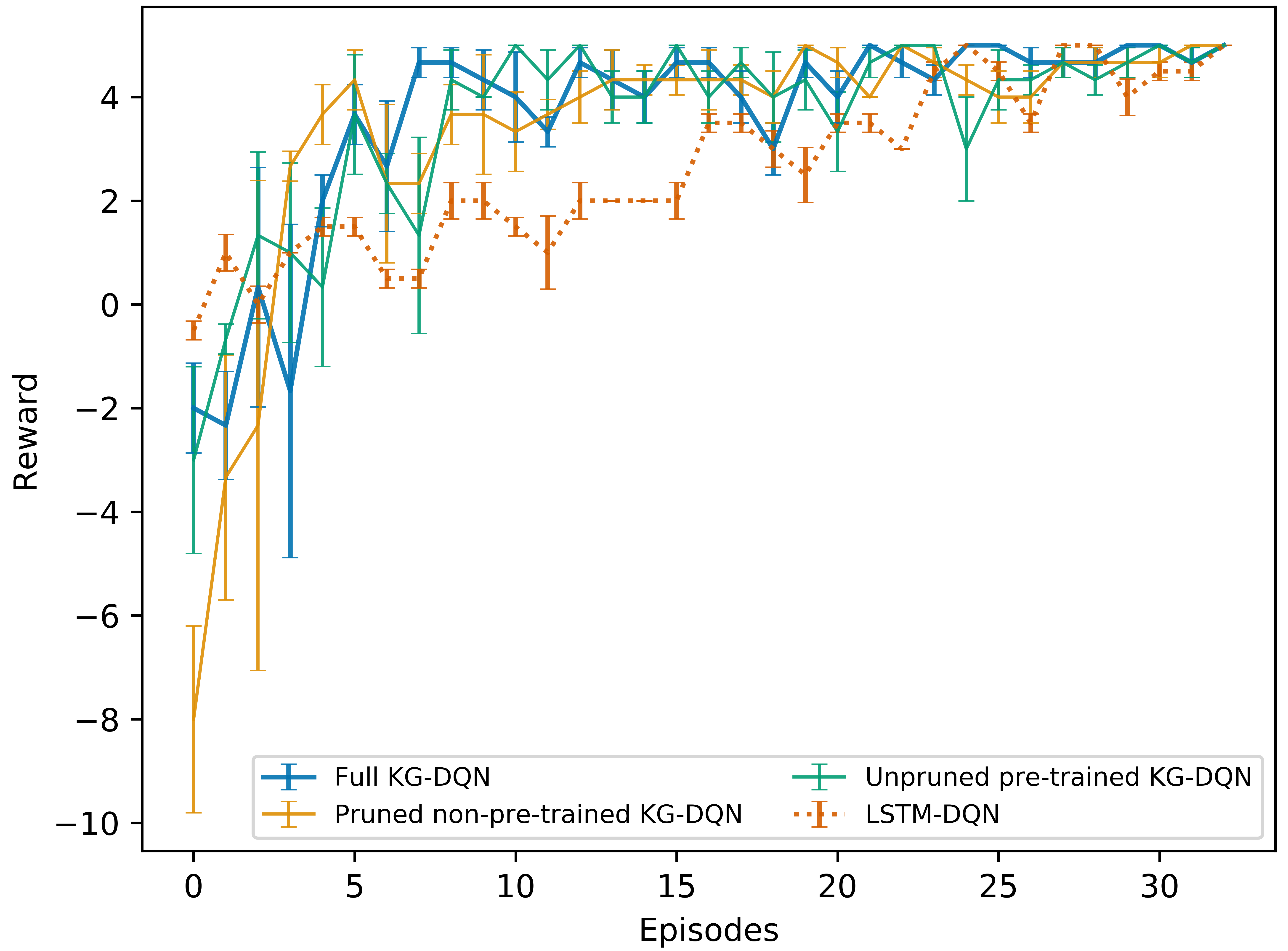}
    \caption{Reward learning curve for select experiments with the small games.}
    \label{fig:reward}
\end{figure}

\begin{figure}[tb]
    \centering
    \includegraphics[width=1\linewidth]{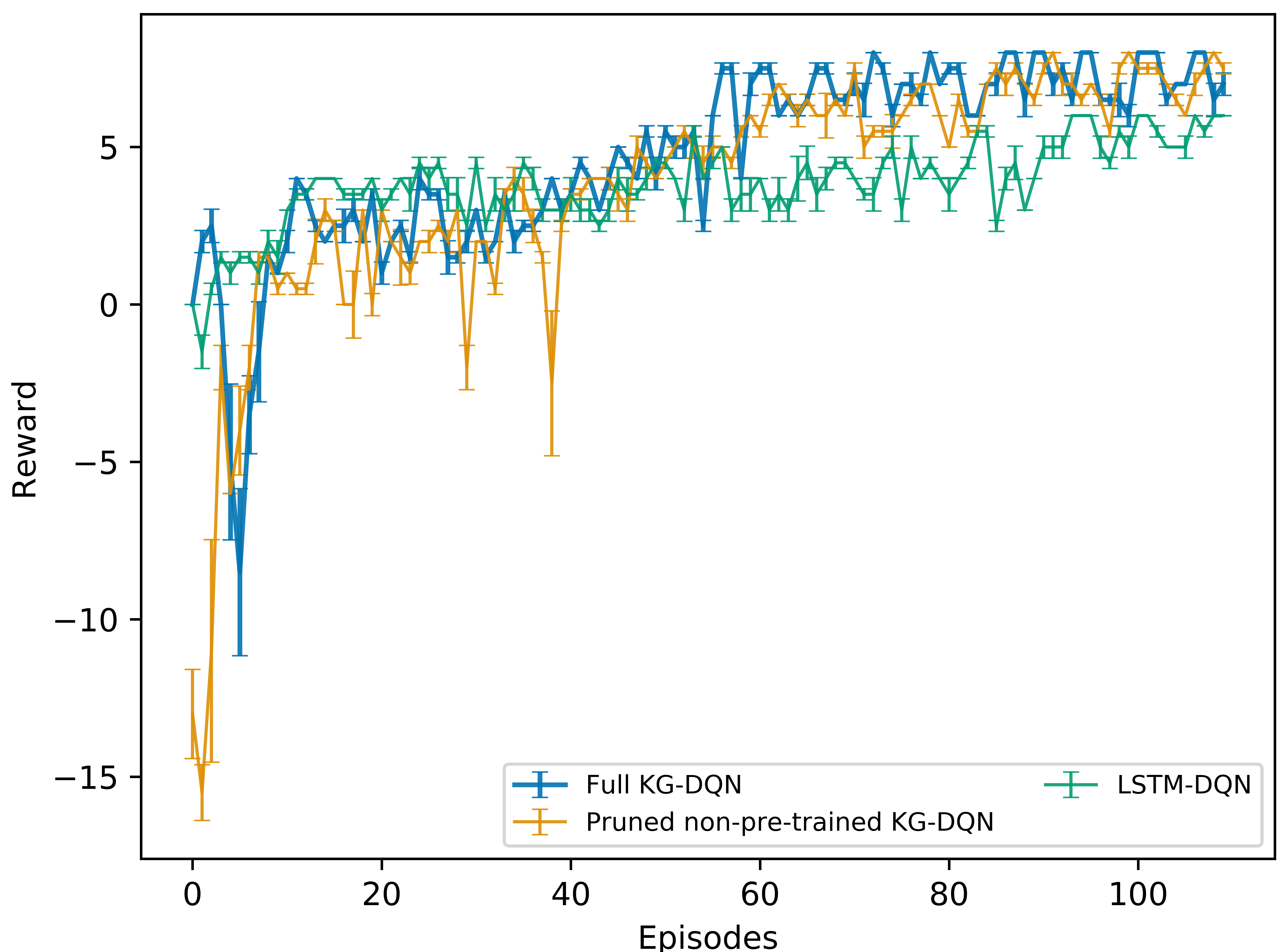}
    \caption{Reward learning curve for select experiments with the large games.}
    \label{fig:reward2}
\end{figure}

On both small and large maps, all versions of KG-DQN tested converge faster than baselines (see Figure~\ref{fig:reward} for the small game and Figure~\ref{fig:reward2} for the large game). 
We don't show BOW-DQN because it is strictly inferior to LSTM-DQN in all situations).
KG-DQN converges $~40\%$ faster than baseline on the small game; both KG-DQN and the LSTM-DQN baseline reaches the maximum reward of five.
On the large game, no agents achieve the maximum reward of 10, and the LSTM-DQN requires more than 300 episodes to converge at the same level as KG-DQN.
%
Since all versions of KG-DQN converge at approximately the same rate, we conclude that the knowledge graph---i.e., persistent memory---is the main factor helping convergence time since it is the common element across all experiments.

\begin{table}[tb]
    \caption{Average number of steps (and standard deviation) taken to complete the small game.}
  \footnotesize
    \centering
    \begin{tabular}{ l l }
     \hline
    {\bf Model} & {\bf Steps} \\ 
      \hline
      Random Command &  319.8\\
      BOW-DQN & $83.1 \pm 8.0$\\
      LSTM-DQN & $72.4 \pm 4.6$\\
      \hline
      Unpruned, pre-trained KG-DQN & $131.7 \pm 7.7$\\
      Pruned, non-pre-trained KG-DQN & $97.3 \pm 9.0$\\
      Full KG-DQN & $73.7 \pm 8.5$\\
      \hline
    \end{tabular}
    \label{table:steps}
  
\end{table}

After training is complete, we measure the number of steps each agent needs to complete each quest.
Full KG-DQN requires an equivalent number of steps in the small game (Table~\ref{table:steps}) and in the large game (Table~\ref{table:steps2}). 
Differences between LSTM-DQN and full KG-DQN are not statistically significant, $p=0.199$ on an independent T-test.
%
The ablated versions of KG-DQN---unpruned KG-DQN and non-pre-trained KG-DQN---require many more steps to complete quests.
TextWorld's reward function allows for a lot of exploration of the environment without penalty so it is possible for a model that has converged on reward to complete quests in as few as five steps or in many hundreds of steps.
From these results, we conclude that the pre-training using our question-answering paradigm is allowing the agent to find a general understanding of how to pick good actions even when the agent has never seen the final test game. 
LSTM-DQN also learns how to choose actions efficiently, but this knowledge is captured in the LSTM's cell state, whereas in KG-DQN this knowledge is made explicit in the knowledge graph and retrieved effectively by graph attention.
Taken together, KG-DQN converges faster without loss of quest solution quality.

\begin{table}[tb]
    \caption{Average number of steps (and standard deviation) taken to complete the large game.}
  \footnotesize
    \centering
    \begin{tabular}{ l l r }
    \hline
    {\bf Model} & {\bf Steps} \\ 
      \hline
      Random Command & 2054.8 \\
      LSTM-DQN & 260.3 $\pm$ 4.5\\
      \hline
      Pruned, non-pre-trained KG-DQN & 340 $\pm$ 6.4\\
      Full KG-DQN & 265.9 $\pm$ 9.4\\
      \hline
    \end{tabular}

    \label{table:steps2}
\end{table}

\section{Conclusions}

We have shown that incorporating knowledge graphs into an deep $Q$-network can reduce training time for agents playing text-adventure games of various lengths. 
We speculate that this is because the knowledge graph provides a persistent memory of the world as it is being explored.
While the knowledge graph allows the agent to reach optimal reward more quickly, it doesn't ensure a high quality solution to quests.
Action pruning using the knowledge graph and pre-training of the embeddings used in the deep $Q$-network result in shorter action sequences needed to complete quests.

The insight into pre-training portions of the agent's architecture is based on converting text-adventure game playing into a question-answering activity.
That is, at every step, the agent is asking---and trying to answer---what is the most important thing to try.
The pre-training acts as a form of transfer learning from different, but related games.
However, question-answering alone cannot solve the text-adventure playing problem because there will always be some trial and error required.

By addressing the challenges of partial observability and combinatorially large action, spaces through persistent memory, our work on playing text-adventure games addresses a critical need for  reinforcement learning for language. 
Text-adventure games can be seen as a stepping stone toward more complex, real-world tasks;
the human world is one of partial understanding through communication and acting on the world using language. 


\bibliography{naaclhlt2019}
\bibliographystyle{acl_natbib}



\end{document}